# Exploiting DINOv3-Based Self-Supervised Features for Robust Few-Shot Medical Image Segmentation

Guoping Xu[1], Jayaram K. Udupa[2], Weiguo Lu[1], You Zhang[1#]

[1]The Medical Artificial Intelligence and Automation (MAIA) Laboratory, Department of Radiation Oncology, University of Texas Southwestern Medical Center, Dallas, TX 75390, USA

[2]Medical Image Processing Group (MIPG), Department of Radiology, University of Pennsylvania, Philadelphia, PA 19104, USA

[#]Email: You.Zhang@UTSouthwestern.edu

## Abstract

Deep learning-based automatic medical image segmentation plays a critical role in clinical diagnosis and treatment planning but remains challenging in few-shot scenarios due to the scarcity of annotated training data. Recently, self-supervised foundation models such as DINOv3, which were trained on large natural image datasets, have shown strong potential for dense feature extraction that can help with the few-shot learning challenge. Yet, their direct application to medical images is hindered by domain differences. In this work, we propose DINO-AugSeg, a novel framework that leverages DINOv3 features to address the few-shot medical image segmentation challenge. Specifically, we introduce WT-Aug, a wavelet-based feature-level augmentation module that enriches the diversity of DINOv3-extracted features by perturbing frequency components, and CG-Fuse, a contextual information-guided fusion module that exploits cross-attention to integrate semantic-rich low-resolution features with spatially detailed high-resolution features. Extensive experiments on six public benchmarks spanning five imaging modalities, including MRI, CT, ultrasound, endoscopy, and dermoscopy, demonstrate that DINO-AugSeg consistently outperforms existing methods under limited-sample conditions. The results highlight the effectiveness of incorporating wavelet-domain augmentation and contextual fusion for robust feature representation, suggesting DINO-AugSeg as a promising direction for advancing few-shot medical image segmentation. Code and data will be made available on https://github.com/apple1986/DINO-AugSeg.

# 1. Introduction

Medical image segmentation aims to assign each pixel (or voxel) of an image to a specific anatomical structure or lesion class. It is a fundamental task in medical image analysis, playing a critical role in diagnosis [1], disease monitoring [2], treatment planning [3], surgical guidance [4], and quantitative assessment of pathology [5]. The advent of deep learning has substantially advanced the field, with U-Net emerging as one of the most influential architectures. It extends the conventional encoder–decoder paradigm by introducing skip connections, which enable the integration of multi-scale contextual and fine-grained features for more precise segmentation [6]. The elegant and efficient design, coupled with its strong performance across diverse imaging modalities, has made it a cornerstone of modern medical image segmentation research [7]. Building on U-Net, a series of variants have been proposed to further enhance performance [8]. For instance, SegResNet employs residual blocks to improve gradient flow and feature representation [9]; UNet++ leverages nested and dense skip connections for more effective multi-scale feature fusion [10]; and nnU-Net introduces a self-adapting, self-configuring framework that automatically optimizes preprocessing, network architecture, training schedules, and postprocessing strategies for a given dataset [11]. Collectively, these advances highlight the continuing evolution of convolution-based U-Net derivatives and their central role in medical image segmentation. Beyond purely convolutional architectures, recent research has increasingly incorporated Transformers into segmentation frameworks, motivated by their capacity to capture long-range dependencies across image regions and thereby address the inherent locality limitations of convolution-based methods. Hybrid approaches such as TransUNet [12] and LeViT-UNet [13] combine convolutional backbones with Transformer modules to balance local feature extraction with global context modeling. In contrast, fully Transformer-based architectures, including UNETR [14] and SwinUNETR [15], leverage self-attention mechanisms to enable end-to-end volumetric segmentation with enhanced contextual representation.

Although these task-specific approaches have achieved significant advances in medical image segmentation, several limitations hinder their translation into routine clinical practice. *A key drawback is the reliance on large, annotated datasets for supervised training*. Constructing such labeled datasets is both costly and labor-intensive, as it requires expert clinicians to perform meticulous slice-by-slice labeling. Inspired by the success of large language models (LLMs) [16, 17], self-supervised learning (SSL) has emerged as a promising alternative by exploiting unlabeled data. This paradigm shift has facilitated the transition from small-scale, task-specific models to large-scale, generalizable frameworks, as curating large collections of unlabeled data is generally more feasible than producing expert annotations.

Following this trajectory, vision foundation models, pre-trained on large-scale natural images using SSL, have demonstrated remarkable generality and robustness as feature extractors for downstream tasks such as object detection and semantic segmentation. Among these, the DINO series represents a prominent line of work [18-20]. In particular, DINOv3 [20], trained on 1.7 billion images with a novel Gram anchoring strategy, exhibits superior capability in extracting high-quality dense feature representations compared to earlier self- and weakly-supervised foundation models, such as DINOv2 [19] and MAE [21], making it especially promising for medical image segmentation. For instance, MedDINOv3 [22] investigated the adaptation of DINOv3 to the medical imaging domain by pre-training on 3.87M axial CT slices under SSL with the same training recipe as DINOv3. Fine-tuning on four segmentation benchmarks, MedDINOv3 achieved performance that matched or surpassed state-of-the-art methods. In addition, recent work [23] further demonstrated that DINOv3 can serve as a strong baseline for extracting robust prior features across medical imaging modalities. As illustrated in Figure 1, DINOv3 demonstrates strong feature robustness under diverse noise corruptions. Principal component analysis (PCA) feature maps derived from the DINOv3 encoder remain consistent across corruption types. Notably, the features extracted by DINOv3 (third row in Figure 1) preserve clear structural delineations of segmented objects, highlighting their reliability under challenging conditions.

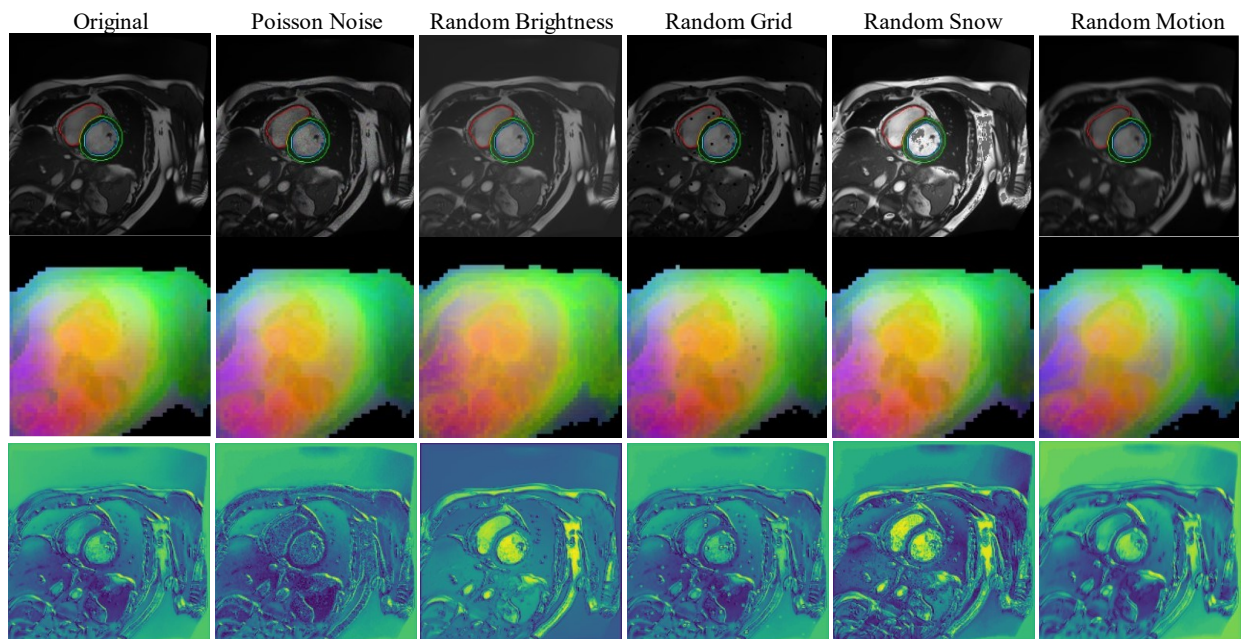

**Figure 1.** The first row presents a representative slice of a cardiac cine MRI set (original) and its variants with different corruption augmentations. Boundaries are overlaid in red, green, and blue, corresponding to the right ventricular cavity, myocardium, and left ventricle, respectively. The second row illustrates principal component analysis (PCA) feature maps extracted from the DINOv3 image encoder. The third row displays feature representations of the first-row images, obtained by applying maximum projection

across the channel dimension in the first stage of DINOv3. These results demonstrate that the features extracted by DINOv3 exhibit strong representational capability and remain robust across diverse corruptions.

Since DINOv3 can extract robust and representative features even under various corruptions, one would expect its stable feature extraction performance will facilitate various downstream image analysis tasks. Indeed, the efficacy of incorporating DINO features has been validated in unsupervised object detection [24] and multimodal image registration [25]. However, when employing DINOv3 as a frozen encoder to combine with a U-Net-like decoder for medical image segmentation of the ACDC dataset [26], we observe a sharp decline in mean Dice score as the number of training samples decreases from 20 to 2, for both the validation and testing sets, as shown in Figure 2. This raises a question: ***why does the segmentation performance degrade markedly with limited training data, even with DINOv3 as a robust feature extractor?*** Given that the DINOv3 encoder is already well pre-trained, we hypothesize that the likely bottleneck lies in the randomly initialized decoder, which usually requires various training samples to robustly learn and adapt to a specific dataset.

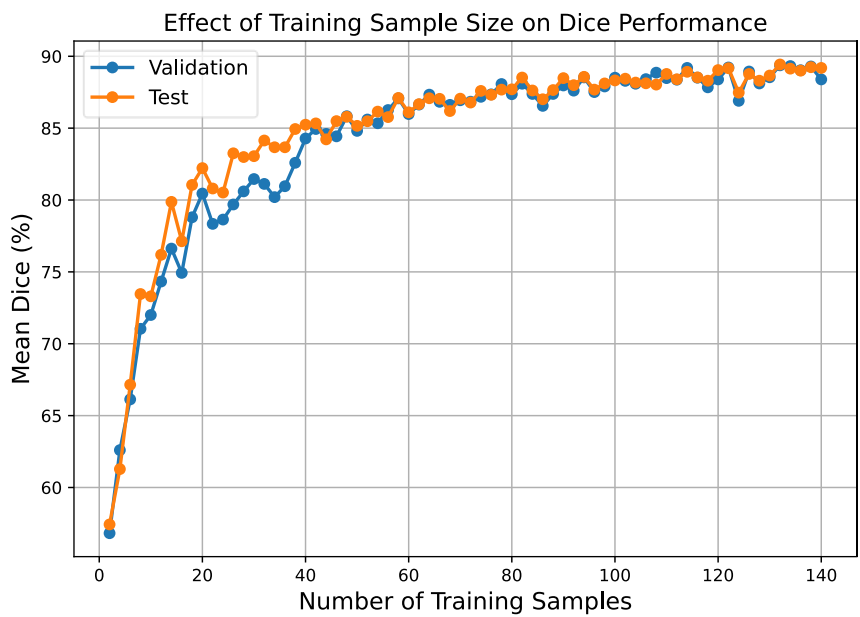

**Figure 2**. Dice similarity scores on the validation and testing sets of the ACDC dataset with varying numbers of training samples (2–140), using the DINOv3 encoder (frozen) and a U-Net-like decoder. A sharp decline in performance is observed when the number of training samples becomes very limited (20 to 2), while the segmentation performance steadily improves as the training sample size increases.

According to our hypothesis, the decoder requires diverse training samples to generalize effectively. A straightforward strategy in few-shot scenarios is to apply image-level augmentation [27]. However, conventional image-level augmentations, such as noise injection, random grid distortions, and brightness adjustments, may provide limited benefit, since the features extracted by DINOv3 remain largely invariant to such perturbations (as illustrated in Figure 1). In this scenario, DINOv3 is functioning as a 'noise filter', which adversely reduces the diversity of the augmented data and thus the generalizability of the trained decoder. The observation motivates this study to investigate whether feature-level augmentation, rather than image-level augmentation, could be more effective in improving the performance of foundational model-assisted, few-shot medical image segmentation under limited labeled data. Compared with image-level augmentation that is performed prior to DINOv3's 'noise-filtration', feature-level augmentation is performed after DIVOv3's feature extraction, thus not affected by the filtration effect. Yet, directly applying classical spatial-domain augmentations to DINO features can compromise feature integrity: for instance, Random Motion may distort object structures (Figure 3, first row), Random Grid may disrupt texture consistency (Figure 3, first and third rows), and Poisson Noise or Random Brightness may obscure critical information or diminish feature discriminability (Figure 3, second row).

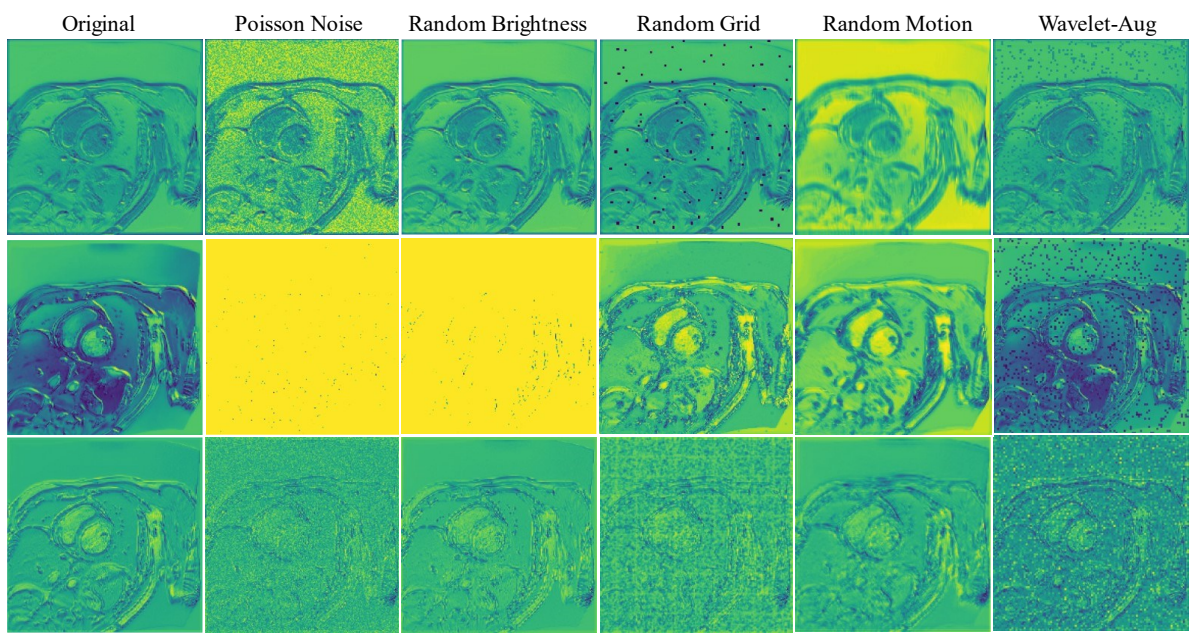

**Figure 3**. Augmentations on features generated at the first stage of the DINOv3 encoder. Rows correspond to the first channel (top), max-pooling across channels (middle), and average-pooling across channels (bottom).

From another perspective, feature-level spatial domain augmentations may substantially alter the feature distribution of the dataset. As illustrated in Figure 4, we applied t-SNE [28] and UMAP [29] to visualize augmented features under different strategies and observed noticeable distribution shifts compared to the original (marked with a star in Figure 4) un-augmented feature distribution. This raises the second critical question: *how can we design feature-level augmentations that avoid drastic distributional shifts* *while still mitigating the challenges of few-shot training in medical image segmentation?* To this end, we investigate feature-level augmentation in the wavelet domain, which enables *independent* manipulation of frequency components. This property offers a favorable trade-off that introduces meaningful variations without severely distorting feature representations, thereby enhancing few-shot segmentation under limited labeled data. Supporting evidence is presented in Figure 3 (last column, wavelet-domain augmentation) and Figure 4 (pink cluster). Specifically, wavelet-domain augmentation preserves structural and textural information (Figure 3) while maintaining feature distributions close to the original samples (Figure 4). In contrast to spatial-domain augmentation, which directly perturbs the entire image or feature map, wavelet-domain augmentation operates independently and randomly on individual wavelet components. As a result, the augmented features better preserve the underlying structural information of the segmented objects.

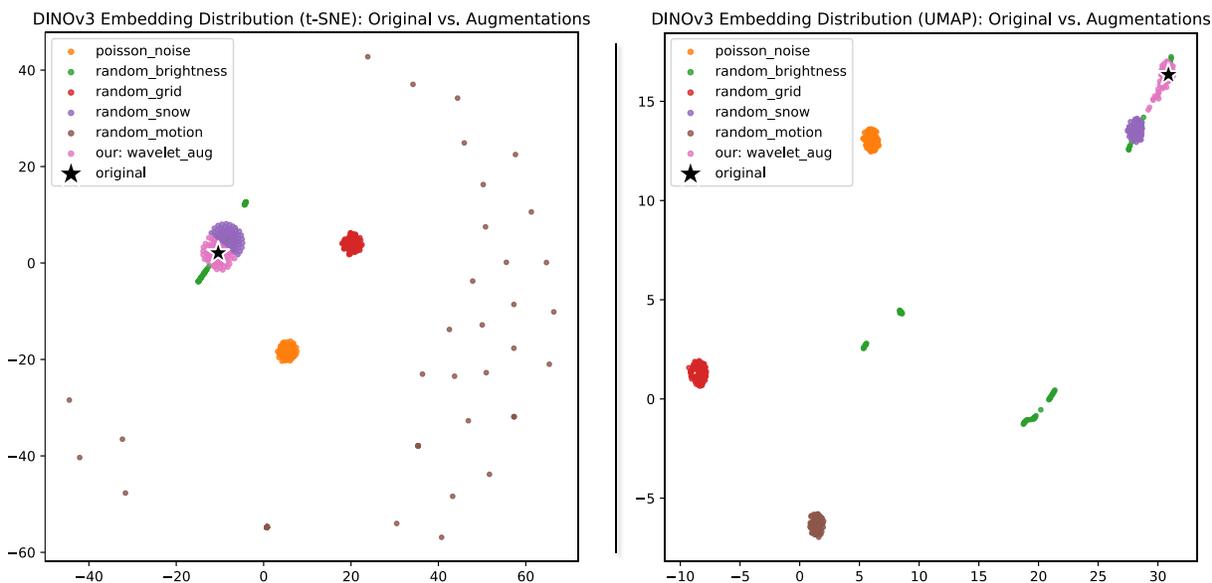

**Figure 4.** Feature distribution of the DINOv3 encoder outputs (global 1D-pooled embeddings) visualized using t-SNE (Left) and UMAP (Right). For simplicity, augmentations are applied directly to the final output embeddings for visualization. In practice, feature-level augmentations are usually performed on the 2D feature maps from different stages of DINOv3 for segmentation.

Beyond feature-level augmentation, an equally critical challenge lies in designing effective feature fusion strategies to supply the decoder with semantically rich and representative information, thereby enhancing the segmentation performance. Classical approaches such as U-Net and nnU-Net [11] fuse encoder and decoder features through direct concatenation, leaving subsequent decoder layers to learn object-specific representations during training. While this strategy is simple and effective, it may not be optimal for fusing DINO-derived features, which already encapsulate high-level contextual semantics [19, 30]. A more principled approach is to leverage these contextual cues to guide the fusion process. To this end, we propose a *contextual-guided feature fusion module (CG-Fuse)* based on a cross-attention mechanism, inspired by the vanilla Transformer [31]. In our design, decoder features [1] are formulated as queries, while encoder features are transformed into keys and values. The underlying motivation is to exploit the rich contextual information embedded in DINO features to guide feature fusion, thereby enabling the decoder to focus more effectively on object-level information critical for accurate segmentation.

In summary, this work introduces two key innovations in leveraging DINOv3-based self-supervised features for robust few-shot medical image segmentation: (**1**) Effective wavelet-domain feature-level augmentation to mitigate the limitations imposed by scarce annotated training samples and the 'noise-filtration' effect of the DINOv3 encoder; and (**2**) The use of rich high-level contextual semantics captured by DINOv3 to guide feature fusion and enhance the decoder's segmentation performance. Based on these two developments, our main contributions are as follows:

(a). We propose a novel *wavelet-based feature-level augmentation method (WT-Aug)* for DINOv3 features.

(b). We design a *contextual-guided feature fusion module (CG-Fuse)* to leverage the high-level contextual information from DINOv3 features.

(c). We introduce a novel segmentation framework, *DINO-AugSeg*, which integrates *WT-Aug* and *CG-Fuse* within an encoder–decoder architecture.

(d). Comprehensive experiments were conducted on six datasets spanning five medical imaging modalities, including MR, CT, ultrasound, endoscopy, and dermoscopy. The results demonstrate the effectiveness of the proposed method, particularly in few-shot segmentation scenarios.

## 2. Related works
(1) Image-level and feature-level augmentation

Data augmentation artificially enlarges the training dataset by generating modified samples from the original limited data, and has proven to be an effective strategy to mitigate the data scarcity challenge in deep learning [32]. Beyond conventional image-level techniques such as brightness adjustment and random rotation, advanced augmentation methods tailored for deep learning have demonstrated remarkable effectiveness [33, 34]. These approaches not only regularize the model but can also be extended to self-supervised learning by encouraging reconstruction of the missing content, thereby driving the network to learn more discriminative features. Building upon this idea, masked autoencoders (MAE) [21] mask random patches of an image and train the model to reconstruct the missing content. MAE has demonstrated that random masking can serve as a scalable and effective pre-training paradigm for large-scale vision models. Inspired by its success, numerous masking-based augmentation strategies have been proposed for self-supervised learning [35-37]. Moreover, such masking-based augmentation strategies have also been explored in the context of medical image segmentation [38-40], where they have shown strong effectiveness in promoting representative feature learning.

---

[1] Here, we designate the features from the final layer of DINOv3 as the queries, as they encapsulate the highest-level semantic information, which is then used to guide the fusion of features from earlier DINOv3 layers.

Beyond image-level augmentation, feature-level augmentation has also been explored as an effective strategy to mitigate the limitations of scarce training data. In [41], a stochastic subsampling approach was introduced to augment features within the encoder block. In [42], feature augmentation was realized by applying random convolutional weights and integrating the resulting features with the original ones through a fixed weighting scheme, thereby facilitating domain-generalized medical image segmentation. UniMatch [43] and UniMatch V2 [44] employed a channel-wise Dropout strategy [45] to perform feature-level augmentation in semi-supervised segmentation frameworks, introducing perturbations that improve the utilization of unlabeled data. Despite their effectiveness, these approaches are confined to the spatial domain, which may inadvertently disrupt the intrinsic distribution of original features (see Figure 4). To address this limitation, we investigate feature-level augmentation in the wavelet domain for few-shot segmentation, leveraging its capacity to decompose features into independent frequency components and thereby introduce controlled variations without severely altering the underlying representation.

(2) Decoder for image segmentation

The decoder in medical image segmentation networks is designed to progressively upsample feature maps to the original resolution, thereby generating the final segmentation masks for each target structure [8]. To enhance prediction quality, it is critical to exploit features from multiple stages of the encoder. Convolution-based architectures such as U-Net, nnU-Net, U-Net++ [10], DeepLabv3+ [46], and PSPNet [47] achieve this through skip connections that fuse encoder and decoder features. While U-Net and nnU-Net primarily concatenate features of identical spatial scales from corresponding encoder-decoder layers, U-Net++ extends this paradigm with densely connected skip pathways that aggregate features across multiple semantic levels. DeepLabv3+ employs dilated convolutions and multi-scale upsampling to capture contextual information, whereas PSPNet integrates pyramid pooling to combine representations at different scales. Hybrid convolution-Transformer approaches, such as TransUNet [12] and LeViT-UNet [13], exploit the complementary strengths of convolutional layers for local feature extraction and Transformers for modeling long-range dependencies, typically by directly concatenating both feature types in the decoder. In addition, pure Transformer-based models, including SegFormer [48] and SegDINO [49], rely on multi-level feature fusion from various encoder stages to generate the final semantic predictions.

Distinct from these prior approaches, which fuse features in the decoder primarily through direct concatenation along skip connections, we propose a contextual information–guided fusion module. This design explicitly leverages the superior semantic representations of DINOv3, enabling the decoder to exploit high-level semantic information more effectively for enhanced segmentation performance.

(3) DINO-series and DINOv3 for medical image segmentation

DINO (self-distillation with no labels) is a self-supervised learning framework trained on Vision Transformers (ViTs) [18]. A key finding is that features extracted from DINO contain richer semantic information for dense prediction tasks compared to features from supervised ViTs or convolutional networks [50]. These features have shown superior transferability to downstream applications, such as object segmentation. Building on this, DINOv2 extended the framework to large-scale foundation model training, relying on extensive curated datasets to learn more robust and generalizable visual representations [19]. More recently, DINOv3 was introduced, incorporating advanced self-supervised strategies in data preparation and optimization. Notably, it proposed a novel objective function, Gram anchoring, to mitigate the degradation of dense feature quality during prolonged training [20]. Empirical results demonstrated that DINOv3 produces high-quality, semantically meaningful dense features, outperforming both self-supervised and weakly-supervised foundation models across diverse vision tasks.

Due to their strong capability in dense feature extraction, DINO-based architectures have been widely adopted in medical image segmentation [49, 51, 52]. For instance, SegDINO leverages a frozen DINOv3 as the encoder backbone combined with a lightweight MLP-based decoder, achieving consistent state-of-the-art performance across multiple medical imaging modalities. Similarly, Dino U-Net [53], designed on an encoder-decoder paradigm, exploits the high-fidelity dense features from DINOv3 to enhance segmentation accuracy. Collectively, these approaches demonstrate that the semantic-rich dense features learned by DINOv3 provide a powerful and generalized foundation for medical image segmentation.

In this study, we investigate the robustness of DINOv3 features for few-shot medical image segmentation. Unlike prior approaches, our focus is two-fold: (1) developing feature augmentation strategies for DINOv3 representations to mitigate the challenges posed by limited training samples, and (2) effectively exploiting contextual information within DINOv3 features to guide feature fusion in the decoder, thereby enhancing the final segmentation performance.

## 3. Method

### 3.1 Overall architecture of DINO-AugSeg

The overall architecture of the proposed DINO-AugSeg framework is illustrated in Figure 5. It is composed of three main components: (1) a frozen DINOv3 encoder as the backbone, (2) skip connections enhanced with WT-Aug blocks, and (3) a decoder integrated with CG-Fuse modules.

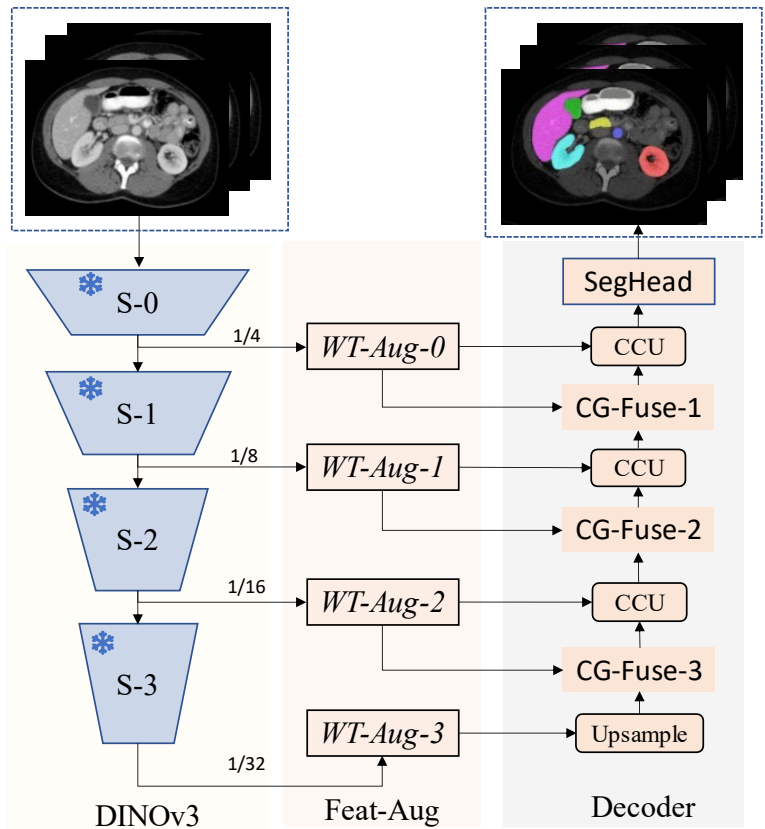

**Figure 5.** Overall architecture of the proposed DINO-AugSeg framework. It comprises a frozen DINOv3 encoder for multi-scale feature extraction, feature-level augmentation blocks (WT-Aug) applied on skip

connections, and a decoder embedded with context-guided fusion (CG-Fuse) modules. CCU denotes the operations of concatenation, convolution, and upsampling via transposed convolution.

First, the input medical images are fed into the frozen DINOv3 encoder, from which multi-scale feature maps are extracted across its four hierarchical stages. These intermediate features are then processed by the corresponding WT-Aug blocks, where feature augmentation is performed in the wavelet domain to enrich the diversity of the representations.

Next, the augmented features are propagated into the decoder. In this stage, the low-resolution but context-rich features from deeper layers are employed to guide the fusion of higher-resolution features through the proposed CG-Fuse module, ensuring effective integration of semantic and spatial information. The output features from CG-Fuse, together with the corresponding features from the encoder at the same stage, are then fed into a CCU module, which performs **c**oncatenation, **c**onvolution, and transposed convolution-based **u**psampling.

Finally, the fused features are passed through a lightweight segmentation head (SegHead, shown in Figure 5) to generate the final segmentation output.

This architectural design aims to leverage the robust semantic features of DINOv3 while introducing feature augmentation and guided fusion strategies to enhance segmentation accuracy and generalization, particularly under limited data conditions.

### 3.2 Wavelet-based feature-level augmentation

The proposed WT-Aug performs feature-level augmentation in the wavelet domain, as illustrated in Figure 6. It consists of three main steps: Haar wavelet[2] decomposition [54], frequency-component augmentation, and reconstruction via the inverse Haar wavelet transform.

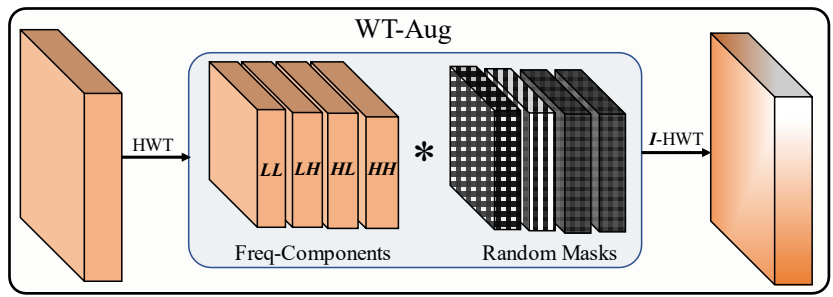

---

[2] We adopt the Haar wavelet transform for its simplicity and computational efficiency compared to other wavelet families (e.g., Daubechies, Symlets, etc.) in this study.

**Figure 6.** Detailed structure of the proposed WT-Aug module. Multi-scale features (shown in brown) extracted from DINOv3 are first decomposed using the Haar wavelet transform into four frequency sub-bands: LL (low-frequency approximation), LH (horizontal high-frequency details), HL (vertical high-frequency details), and HH (diagonal high-frequency details). Each sub-band is then element-wise multiplied with four randomly generated masks to perform feature-level augmentation. Finally, the modified sub-bands are reconstructed into the spatial feature domain through the inverse Haar wavelet transform.

Specifically, the feature maps from one stage of DINOv3 output are first decomposed by Haar wavelet transform into four independent frequency sub-bands: LL (low-frequency approximation), LH (horizontal high-frequency details), HL (vertical high-frequency details), and HH (diagonal high-frequency details). This process can be formulated as follows:

$$(LL, LH, HL, HH) = WT(F), \qquad (1)$$

where $F$ denotes the input feature map and $WT(.)$ represents the Haar wavelet transform.

Next, each frequency sub-band is element-wise multiplied with a randomly generated mask of the same spatial dimension to produce augmented frequency components:

$$(LL', LH', HL', HH') = (LL \odot M_{LL}, LH \odot M_{LH}, HL \odot M_{HL}, HH \odot M_{HH}), \qquad (2)$$

where $\odot$ denotes pixel-wise multiplication and $M_*$ represents the random masks.

Finally, the augmented frequency components are reconstructed into the spatial domain through the inverse Haar wavelet transform:

$$F' = WT^{-1}(LL', LH', HL', HH'), \qquad (3)$$

This augmentation strategy effectively introduces stochastic perturbations by selectively altering a portion of the frequency information while preserving the structural and textural integrity of objects. As a result, the reconstructed features remain semantically consistent but diverse in intensities, thereby helping to enhance robustness and generalization under limited training data conditions.

### 3.3 Contextual-guided feature fusion module

The detailed structure of the proposed CG-Fuse module is illustrated in Figure 7(a). Specifically, the low-resolution but context-rich features from the deeper decoder layers are first upsampled (directly for the deepest layer, or through CCU for the shallower layers) and projected through a linear layer to form the query ($Q$). In parallel, the augmented features from the WT-Aug module one-level up are projected to serve as the key ($K$) and value ($V$), respectively. The triplet ($Q$, $K$, $V$) is then fed into a multi-head cross-attention mechanism (Figure 7(b)). For each attention head, the attention map is computed as:

$$Cross\text{-}Attention(Q, K, V) = Softmax\left(\frac{QK^T}{\sqrt{d_K}}\right)V, \qquad (4)$$

where $d_K$ denotes the dimension of the key. The softmax-normalized attention weights ensure stable learning and emphasize the most relevant contextual information (Figure 7(c)). Finally, the outputs from all attention heads are concatenated and passed through a feed-forward layer. A residual skip connection is applied by adding the original decoder features, ensuring stable training and effective feature fusion.

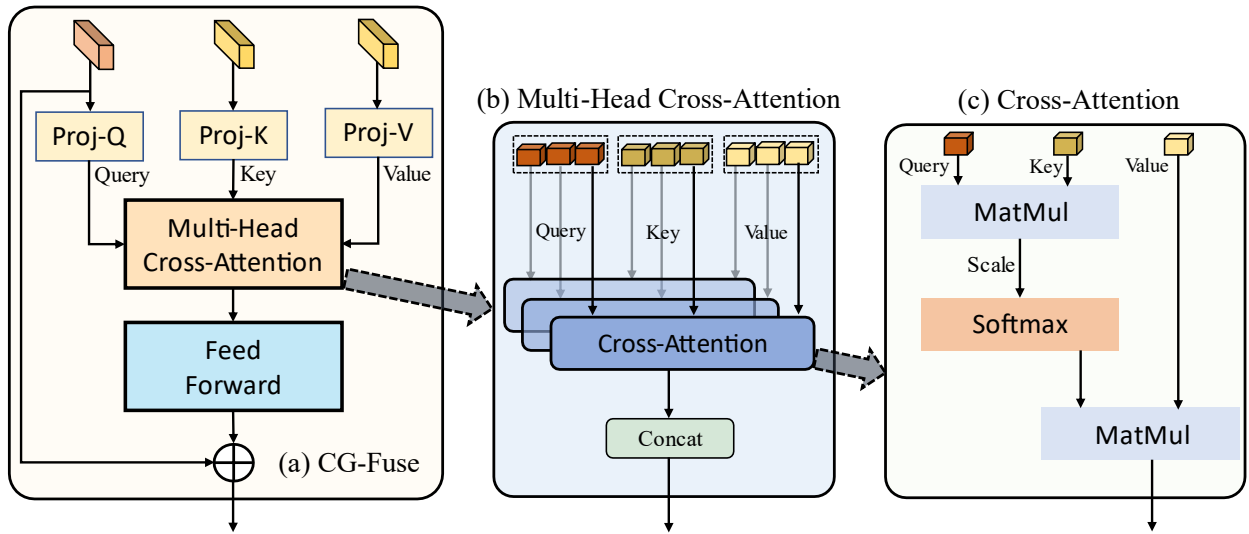

**Figure 7.** The structure of (a) CG-Fuse, (b) Multi-Head Cross-Attention, and (c) Cross-Attention.

## 4. Experiments

### 4.1 Datasets

(1) ACDC dataset

The <u>A</u>utomated <u>C</u>ardiac <u>D</u>iagnosis <u>C</u>hallenge (ACDC) dataset comprises cardiac cine-MRI scans, with each case containing two phases: end-systolic and end-diastolic. All scans are manually annotated for the left ventricle (LV), myocardium (MYO), and right ventricle (RV), serving as 'ground truth'. Following the data splits in [13], the same 20 cases (40 3D scans) are used for testing, while the remaining 80 cases (160 scans) are used for training.

(2) LA2018 dataset

The Left Atrium Segmentation Challenge (LA2018) dataset [55] contains 100 3D gadolinium-enhanced MR imaging scans (GE-MRIs) and corresponding LA segmentation masks. Following the data splitting in [56], we use the same 20 scans for testing, while the remaining 80 scans are used for training.

(3) Synapse dataset

The Synapse Multi-Organ Segmentation (Synapse) dataset comprises 30 abdominal CT scans with annotations for multiple organs, including the aorta, gallbladder, spleen, left kidney, right kidney, liver, pancreas, and stomach. Following the data split in [13], 12 scans are reserved for testing, while the remaining 18 scans are used for training.

(4) TN3K dataset

The TN3K dataset [57] is a publicly available collection for research on thyroid nodule segmentation from ultrasound images. It contains 3,493 images from 2,421 patients, with 614 images reserved for testing. The remaining 2,879 images are used for training.

(5) Kvasir-SEG dataset

The Kvasir-SEG dataset [58] contains 1,000 endoscopic images of polyps with corresponding 'ground-truth' annotations, along with an additional testing set without public labels. In [58], the dataset was divided into 880 training images and 120 validation images. In this study, we used the 880 images as the training dataset, while using the 120 images for testing.

(6) ISIC2018 dataset

The ISIC2018 dataset [59] is part of the International Skin Imaging Collaboration (ISIC) challenges and is widely used for skin lesion segmentation. It consists of 2,594 dermoscopic images with corresponding expert-annotated lesion masks. Following the official challenge split, 2,594 images are provided for training and validation, while an independent test set is maintained by the organizers without released 'ground-truth' labels. In this study, we randomly divide the labeled 2,594 images into 2,074 for training and 520 for testing.

Table 1 summarizes the characteristics of the six benchmark datasets used in our study, including their imaging modality, annotated structures, and the number of cases/images for training and testing in each split. Note that the training numbers only show the maximum scans/images assigned to each set that allow for training. To simulate few-shot learning scenarios, we only randomly selected a few cases from them for model training (detailed in the results section). Regardless of the scan/image numbers used in few-shot training, all the samples in the testing split were used in model evaluation.

**Table 1.** Summary of the six public datasets used for evaluation.

| Dataset | Modality | Objects | Train | Test |
|---|---|---|---|---|
| ACDC | MR | Cardiac | 160 | 40 |
| LA2018 | MR | Left Atrium | 80 | 20 |
| Synapse | CT | Abdominal Organ | 18 | 12 |
| TN3K | US | Thyroid | 2,879 | 614 |
| Kvasir-SEG | Endoscopy | Polyp | 880 | 120 |
| ISIC2018 | Dermoscopy | Skin Lesion | 2,074 | 520 |

4.2 Implementation details

To comprehensively evaluate the effectiveness of DINO-AugSeg, we compare it with representative state-of-the-art segmentation models across both convolutional and transformer paradigms. The convolutional baselines include Attention-Unet [60], MultiResUNet [61], nnU-Net [11], SegNet [62], SegResNet [63], U-Net [7], and Unet++ [10], while the transformer-based models include MISSFormer [64], SegFormer [48], and SwinUNETR [15]. We further benchmark DINO-AugSeg against leading self-supervised representation learning methods, including AIMv2 [65], MAE [21], MoCov2 [66], and SimSiam [67], which use the self-supervised features as frozen encoders within U-Net-style architectures. Additionally, we compare our approach with the recently proposed SegDINO [49], which also freezes the DINOv3 encoder but uses a lightweight MLP-based decoder.

All experiments are implemented using the PyTorch framework and executed on an NVIDIA RTX 4090 GPU running Ubuntu 22.04.5 LTS. For a fair comparison, all models are optimized with Adam, using a learning rate of $1\times10^{-4}$. The training objective combines cross-entropy and Dice loss with the same weights. For SegDINO and our proposed DINO-AugSeg, the DINOv3 backbone is frozen and only the segmentation decoder is trained. Similarly, for other self-supervised comparison methods (AIMv2, MAE, MoCov2, and SimSiam), the feature extraction encoders are frozen, whereas for all other models the entire network is trained end-to-end. The experimental configurations, including input image size, batch size, and training epochs for each dataset, are summarized in Table 2. All models are trained and tested in 2D. For DINO-AugSeg, the WT-Aug module is applied only during training and is not used during the testing stage. For evaluation, we adopt the Dice similarity coefficient (DICE) and the 95th percentile Hausdorff distance (HD95). Specifically, Dice and HD95 are computed on 3D volumes for ACDC, LA2018, and Synapse, while for TN3K, Kvasir-SEG, and ISIC2018 they are calculated on 2D images. For datasets with multiple segmentation targets (for instance, ACDC or Synapse), the corresponding metrics were averaged across the segmentation targets.

**Table 2.** Experimental settings for each dataset, including input image size, batch size, and training epochs used.

| Dataset | Image Size | Batch Size | Epochs |
|---|---|---|---|
| ACDC | 768×768 | 2 | 2000 |
| LA2018 | 768×768 | 2 | 1000 |
| Synapse | 224×224 | 2 | 500 |
| TN3K | 512×512 | 4 | 500 |
| Kvasir-SEG | 512×512 | 2 | 1000 |
| ISIC2018 | 224×224 | 4 | 250 |

### 4.3 Results

4.3.1 Performance of few-shot learning

(1) Few-shot segmentation results on the ACDC, LA2018, and Synapse datasets (volumetric image)

We first evaluated the segmentation performance of DINO-AugSeg under the 'one-shot' training scenario, where only a single annotated scan was used for training each model. Table 3 summarizes the results on three volumetric datasets: ACDC, LA2018, and Synapse, comparing our method with a range of state-of-the-art approaches, including convolution-based and transformer-based architectures, as well as recent self-supervised learning methods. As shown, DINO-AugSeg achieves the highest Dice scores and lowest HD95 values on the ACDC and LA2018 datasets, and delivers competitive performance on Synapse. These results demonstrate that DINO-AugSeg possesses strong generalization ability and robustness, even in extremely low-data regimes.

**Table 3.** Experimental results on ACDC, LA2018, and Synapse datasets with one training scan. Arrows are pointing in the direction of improved accuracy. The bold metric results indicate the best values.

| Dataset | | ACDC | | LA2018 | | Synapse | |
|---|---|---|---|---|---|---|---|
| Method | Type | DICE (%)↑ | HD95 (vol)↓ | DICE (%)↑ | HD95 (vol)↓ | DICE (%)↑ | HD95 (vol)↓ |
| Attention-Unet | Conv | 0.73 | 106.65 | 23.73 | 44.48 | 39.93 | 87.19 |
| MultiResUNet | Conv | 9.41 | 82.46 | 21.85 | 53.70 | 0.57 | 196.07 |
| nn-UNet | Conv | 22.45 | 91.13 | 55.03 | 40.56 | 23.30 | 183.46 |
| SegNet | Conv | 22.26 | 64.35 | 37.46 | 29.37 | 17.09 | 87.25 |
| SegResNet | Conv | 22.84 | 76.08 | 29.10 | 45.07 | 29.20 | 121.51 |
| U-Net | Conv | 25.89 | 58.39 | 20.25 | 44.83 | 42.08 | 97.74 |

| Method | Encoder Type | DICE (%)↑ | HD95 (vol)↓ | DICE (%)↑ | HD95 (vol)↓ | DICE (%)↑ | HD95 (vol)↓ |
|---|---|---|---|---|---|---|---|
| Unet++ | Conv | 22.86 | 64.08 | 24.96 | 41.25 | 44.56 | 108.02 |
| MISSFormer | Trans | 23.39 | 54.24 | 32.46 | 40.48 | 20.15 | 87.10 |
| SegFormer | Trans | 17.16 | 75.81 | 36.72 | 27.04 | 15.48 | 90.68 |
| SwinUNETR | Trans | 25.61 | 63.58 | 35.34 | 33.62 | 24.65 | 129.46 |
| AIMv2 | SSL | 47.29 | 40.32 | 49.25 | 24.72 | 33.62 | 94.36 |
| MAE | SSL | 41.63 | 18.27 | 47.79 | 28.17 | 3.34 | 118.75 |
| MoCov2 | SSL | 60.64 | 44.84 | 39.04 | 36.63 | 44.71 | 85.17 |
| SimSiam | SSL | 49.32 | 41.31 | 22.72 | 32.38 | **45.02** | 93.29 |
| SegDINO | SSL | 60.39 | 26.39 | 48.45 | 34.20 | 20.88 | **68.32** |
| DINO-AugSeg | SSL | **71.70** | **23.69** | **75.43** | **22.06** | 42.84 | 93.47 |

We further assessed the performance under the 'seven-shot' training scenarios, where seven annotated scans were used for training. Table 4 presents the results on the same three datasets. Again, DINO-AugSeg consistently achieves the highest Dice scores across all datasets while maintaining competitive HD95 values. These results demonstrate that DINO-AugSeg effectively leverages limited annotated data to achieve superior segmentation performance compared with both traditional and self-supervised baselines. In particular, the Dice scores improved by 10.15%, 10.79%, and 28.35% (in absolute terms, same in the following) on the three datasets, respectively, comparing to the 'one-shot' results in Table 3. In addition, the results obtained using two training scans ('two-shot') on these datasets are presented in the supplementary material, which follows the same performance improvement trend.

**Table 4.** Experimental results on ACDC, LA2018, and Synapse datasets with seven training scans. Arrows are pointing in the direction of improved accuracy. The bold metric results indicate the best values.

| Dataset | | ACDC | | LA2018 | | Synapse | |
|---|---|---|---|---|---|---|---|
| Method | Encoder Type | DICE (%)↑ | HD95 (vol)↓ | DICE (%)↑ | HD95 (vol)↓ | DICE (%)↑ | HD95 (vol)↓ |
| Attention-Unet | Conv | 61.98 | 30.59 | 56.48 | 33.10 | 69.76 | 24.44 |
| MultiResUNet | Conv | 26.46 | 55.72 | 48.53 | 39.47 | 5.33 | 250.27 |
| nn-UNet | Conv | 52.09 | 49.57 | 81.54 | 12.58 | 63.11 | 30.55 |
| SegNet | Conv | 54.75 | 31.42 | 61.41 | 20.90 | 52.97 | 29.13 |
| SegResNet | Conv | 54.02 | 52.81 | 63.04 | 28.16 | 67.57 | 26.64 |
| U-Net | Conv | 60.72 | 27.51 | 53.45 | 33.74 | 66.02 | 25.03 |
| Unet++ | Conv | 62.79 | 34.87 | 64.47 | 30.85 | 70.85 | 19.47 |
| MISSFormer | Trans | 41.04 | 20.87 | 76.38 | 18.04 | 51.95 | 34.94 |
| SegFormer | Trans | 39.87 | 43.60 | 65.11 | 23.40 | 43.51 | 41.11 |
| SwinUNETR | Trans | 48.49 | 24.74 | 74.73 | 19.80 | 53.10 | 33.55 |
| AIMv2 | SSL | 61.23 | 18.75 | 77.56 | 16.74 | 57.23 | 19.75 |
| MAE | SSL | 47.99 | 15.04 | 73.40 | 21.68 | 48.48 | 32.19 |
| MoCov2 | SSL | 73.23 | 10.35 | 62.90 | 22.95 | 66.91 | 19.00 |
| SimSiam | SSL | 75.07 | 12.01 | 57.63 | 25.88 | 66.44 | **17.60** |
| SegDINO | SSL | 73.96 | **6.18** | 83.58 | 12.48 | 54.10 | 22.44 |
| DINO-AugSeg | SSL | **81.85** | 9.96 | **86.22** | **11.75** | **71.19** | 21.76 |

(2) Few-shot segmentation results on the TN3K, Kvasir-SEG, and ISIC2018 datasets (2D image)

We further evaluated the few-shot segmentation performance of DINO-AugSeg on three 2D image datasets: TN3K, Kvasir-SEG, and ISIC2018, which cover ultrasound, endoscopy, and dermoscopy modalities, respectively. In this experiment, 25/100, 10/40, and 5/25 annotated the numbers of 'few-shot' samples used for training on TN3K, Kvasir-SEG, and ISIC2018, respectively. Table 5 summarizes the DICE results of Dino-AugSeg, compared with various convolution-based, transformer-based, and self-supervised learning methods.

As shown, DINO-AugSeg achieves the highest Dice scores on TN3K (25 training samples), and Kvasir-SEG under both few-shot settings, substantially surpassing traditional convolutional and transformer-based baselines. On TN3K (100 training samples), SegDINO attains the best performance, while DINO-AugSeg remains a close second. For ISIC2018, although DINO-AugSeg performs slightly below SimSiam and MoCov2, it still achieves competitive results among SSL-based models. Overall, these findings underscore the strong adaptability and generalization capacity of DINO-AugSeg across heterogeneous imaging modalities and varying data scales in few-shot segmentation scenarios.

**Table 5.** DICE performance on TN3K, Kvasir-SEG, and ISIC2018 image datasets with 25/100, 10/40, and 5/25 training samples, respectively. Arrows are pointing in the direction of improved accuracy. The bold metric results indicate the best values.

| Dataset | | TN3K | | Kvasir-SEG | | ISIC2018 | |
|---|---|---|---|---|---|---|---|
| Training Sample Number | | 25 | 100 | 10 | 40 | 5 | 25 |
| Method | Encoder Type | DICE (%)↑ | DICE (%)↑ | DICE (%)↑ | DICE (%)↑ | DICE (%)↑ | DICE (%)↑ |
| Attention-Unet | Conv | 48.01 | 45.65 | 32.78 | 42.59 | 57.40 | 76.98 |
| MultiResUNet | Conv | 30.15 | 37.35 | 31.85 | 34.14 | 61.79 | 72.06 |
| nn-UNet | Conv | 43.40 | 45.46 | 38.24 | 30.08 | 65.90 | 70.33 |
| SegNet | Conv | 44.18 | 55.04 | 32.34 | 39.09 | 54.69 | 71.93 |
| SegResNet | Conv | 42.97 | 51.75 | 31.10 | 29.91 | 62.68 | 74.12 |
| U-Net | Conv | 46.04 | 45.46 | 30.84 | 45.31 | 56.52 | 78.94 |
| Unet++ | Conv | 47.66 | 55.04 | 17.34 | 64.73 | 53.58 | 76.73 |
| MISSFormer | Trans | 23.09 | 47.76 | 16.88 | 25.56 | 61.85 | 75.72 |
| SegFormer | Trans | 35.05 | 45.46 | 15.41 | 39.66 | 67.74 | 77.53 |
| SwinUNETR | Trans | 40.76 | 46.73 | 39.92 | 23.11 | 62.95 | 73.04 |
| AIMv2 | SSL | 53.08 | 50.56 | 57.22 | 54.90 | 66.73 | 74.44 |
| MAE | SSL | 49.45 | 52.99 | 42.69 | 66.96 | 69.99 | 74.06 |
| MoCov2 | SSL | 56.08 | 62.78 | 60.63 | 63.84 | 68.60 | 81.27 |
| SimSiam | SSL | 56.09 | 60.80 | 57.10 | 44.94 | **73.02** | **81.72** |
| SegDINO | SSL | 30.06 | **66.16** | 64.38 | 42.60 | 64.85 | 79.38 |
| DINO-AugSeg | SSL | **60.13** | 65.40 | **73.59** | **78.61** | 67.64 | 78.53 |

### 4.3.2 Ablation study

#### (1) Impacts of various augmentation strategies

We conducted an ablation study on the ACDC dataset to evaluate the impact of different augmentation strategies within the proposed DINO-AugSeg framework. Three categories of augmentation were investigated: image-level, feature-level (spatial-domain), and feature-level (wavelet-domain) augmentation.

For both image-level and feature-level (spatial-domain) augmentation, four commonly used intensity-based transformations were randomly applied on the spatial domain of images or features during training: brightness adjustment, motion blur, Poisson noise, and random pixel masking. For feature-level (wavelet-domain) augmentation, only random pixel masking was applied to the DINOv3 feature maps in the wavelet domain, as described in the Methods section. This simplified design allows us to isolate and clearly assess the contribution of wavelet-domain augmentation to the final segmentation performance.

As summarized in Table 6, feature-level augmentation strategies generally improve segmentation accuracy over image-level augmentation, with wavelet-domain augmentation showing the most consistent gains across different cardiac structures in terms of DICE. We observe that HD95 does not always achieve the best performance for feature-level augmentation in the few-shot setting, regardless of whether augmentation is applied in the spatial or wavelet domain within the DINO-AugSeg framework. We speculate that directly augmenting DINOv3 feature maps may introduce additional noise, making it more challenging for the model to precisely capture fine object boundaries, which in turn affects the HD95 metric.

**Table 6.** Ablation study on the ACDC dataset with different augmentation strategies.

| Augmentation | Domain | 1 Training Sample | | 2 Training Samples | | 7 Training Samples | | All Training Samples | |
|---|---|---|---|---|---|---|---|---|---|
| | | DICE (%)↑ | HD95 (vol)↓ | DICE (%)↑ | HD95 (vol)↓ | DICE (%)↑ | HD95 (vol)↓ | DICE (%)↑ | HD95 (vol)↓ |
| Image-Level | Spatial | 67.80 | **17.95** | 68.68 | 13.86 | 75.16 | **5.80** | 86.08 | 3.21 |
| Feature-Level | Spatial | 68.65 | 19.47 | 70.52 | **6.71** | 80.38 | 8.53 | 90.38 | 1.75 |
| | Wavelet | **71.70** | 23.69 | **76.24** | 15.56 | **81.85** | 9.96 | **91.29** | **1.66** |

#### (2) Comparison Between Different Decoders Using DINOv3 as the Encoder

To further assess the effectiveness of the proposed CG-Fuse decoder, we compared it with six widely adopted decoder architectures: DeepLabv3+ [46], OCRNet [68], PSPNet [47], SegFormer, U-Net, and UNet++, while keeping DINOv3 as the shared encoder. The experiments were conducted on the ACDC and LA2018 datasets under the seven-shot training setting, where only seven annotated scans were available for training. As presented in Table 7, the proposed CG-Fuse decoder consistently achieved superior segmentation performance across both datasets. It attained the highest Dice scores of 81.85% and 86.22%, along with competitive HD95 values of 9.96 and 11.75, on the ACDC and LA2018 datasets, respectively. These results demonstrate that the CG-Fuse decoder effectively enhances multi-scale feature integration and spatial coherence, outperforming conventional decoders in both convolutional and transformer-based frameworks.

**Table 7.** Comparison between different decoder architectures for DINO-AugSeg, on the ACDC and LA2018 datasets under the seven-shot training setting. Arrows are pointing in the direction of improved accuracy. The bold metric results indicate the best values.

| Dataset | ACDC | | LA2018 | |
|---|---|---|---|---|
| Decoder | DICE (%)↑ | HD95 (vol)↓ | DICE (%)↑ | HD95 (vol)↓ |
| Deeplabv3plus | 63.63 | 9.85 | 73.79 | 20.85 |
| OCRNet | 73.41 | 10.08 | 72.89 | 12.85 |
| PSPNet | 70.93 | **5.07** | 71.98 | 13.97 |
| SegFormer | 77.68 | 6.98 | 81.51 | 15.89 |
| U-Net | 72.81 | 10.24 | 84.62 | 13.12 |
| UNet++ | 79.34 | 6.71 | 84.13 | 15.55 |
| CG-Fuse (Our) | **81.85** | 9.96 | **86.22** | **11.75** |

(3) Effects of WT-Aug and CG-Fuse under the Seven-Shot Training Setting

To further investigate the contributions of the WT-Aug module and the proposed CG-Fuse decoder, we conducted an ablation study under the seven-shot training setting on the ACDC and LA2018 datasets. In this experiment, we compared four configurations: with and without WT-Aug, and with and without CG-Fuse (where the latter is replaced by a standard U-Net-like decoder that directly concatenates encoder and decoder features).

As shown in Table 8, both WT-Aug and CG-Fuse individually improved segmentation performance, and their combination yielded the best overall results. Specifically, incorporating both modules achieved the highest Dice scores of 81.85% and 86.22%, along with the lowest HD95 values of 9.96 and 11.75, on the ACDC and LA2018 datasets, respectively. These findings confirm that WT-Aug enhances data and feature diversity, while CG-Fuse strengthens feature interaction and spatial consistency in the decoding stage.

**Table 8.** Effects of WT-Aug and CG-Fuse on DINO-AugSeg under the seven-shot training setting on ACDC and LA2018 datasets. Arrows are pointing in the direction of improved accuracy. The bold metric results indicate the best values.

| Dataset | | ACDC | | LA2018 | |
|---|---|---|---|---|---|
| WT-Aug | CG-Fuse | DICE (%)↑ | HD95 (vol)↓ | DICE (%)↑ | HD95 (vol)↓ |
| ✓ | ✓ | **81.85** | **9.96** | **86.22** | **11.75** |
| × | ✓ | 76.98 | 16.87 | 85.61 | 13.79 |
| ✓ | × | 77.32 | 11.82 | 85.05 | 16.23 |
| × | × | 72.81 | 10.24 | 84.62 | 13.12 |

(4). Influence of image size on segmentation performance

We observed that DINO-AugSeg did not achieve state-of-the-art performance across all datasets, particularly on Synapse/ISIC2018, where several SSL-based methods such as SimSiam and MoCov2 performed better. We hypothesize that this limitation may arise from the downscaled input representation in DINOv3, which was originally trained on large matrix-size, high-resolution images. However, for small image-size datasets like Synapse/ISIC2018 (Table 2), downscaling the input excessively leads to the loss of fine-grained spatial details critical for precise boundary delineation. To validate the hypothesis, we performed another ablation study by upsampling the Synapse/ISIC2018 datasets to higher resolutions (512×512) to offset the downscaling effects of the DINOv3 encoder.

**Table 9.** Effect of different input image sizes on the performance of SimSiam and DINO-AugSeg under varying amounts of training data (1, 2, and 7 scans for Synapse; 5, 10, and 25 images for ISIC2018), reported in terms of the Dice score.

| Method | Dataset | Synapse | | | ISIC2018 | | |
|---|---|---|---|---|---|---|---|
| Name | Image size | 1 | 2 | 7 | 5 | 10 | 25 |
| SimSiam | 224×224 | 45.02 | 47.00 | 66.44 | 73.02 | 78.12 | 81.72 |
| SimSiam | 512×512 | 50.11 | 55.60 | 69.56 | 72.35 | 73.90 | 77.70 |
| DINO-AugSeg | 224×224 | 42.84 | 50.37 | 71.19 | 67.64 | 74.09 | 78.53 |
| DINO-AugSeg | 512×512 | **54.65** | **63.92** | **77.24** | **77.40** | **81.14** | **83.81** |

Table 9 summarizes the impact of different input image sizes on the performance of SimSiam and DINO-AugSeg across varying numbers of training samples (1, 2, and 7 scans for Synapse, and 5, 10, and 25 images for ISIC2018). As shown, increasing the input resolution from 224×224 to 512×512 consistently improves the Dice scores for both methods on Synapse, with DINO-AugSeg exhibiting a notably larger performance gain. For ISIC2018, we observe that DINO-AugSeg achieves substantial improvements at higher resolutions and outperforms SimSiam. This indicates that by using a larger input matrix, the downscaling effect of DINOv3 encoders in DINO-AugSeg is reduced, which helps to preserve the fine-grained details towards accurate image segmentation. In contrast, SimSiam experiences a performance drop on the ISIC2018 dataset when increasing the input size from 224×224 to 512×512. This is likely due to the fact that SimSiam uses a ResNet-50 encoder pre-trained on natural images at a matrix size of 224×224, rendering the original 224x224 input size more appropriate. However, for the CT-based Synapse dataset, SimSiam also improves with a larger input size, suggesting complex interplay between medical image characteristics and matrix size.

4.3.3 Visual Comparison

To qualitatively compare the segmentation performance, Figure 8 presents representative results on the ACDC, LA2018, and Synapse datasets across MR (first and second rows) and CT (third row) modalities under the 7-shot training setting. The first column shows the ground-truth masks overlaid on the original images, while the remaining columns display the prediction overlays produced by DINO-AugSeg, nn-UNet, SwinUNETR, SimSiam, and SegDINO, representing state-of-the-art convolution-based (nn-UNet), Transformer-based (SwinUNETR), and self-supervised learning methods (DINO-AugSeg, SimSiam, and SegDINO). As observed, DINO-AugSeg demonstrates robust and stable segmentation performance across diverse anatomical structures in both MR and CT images.

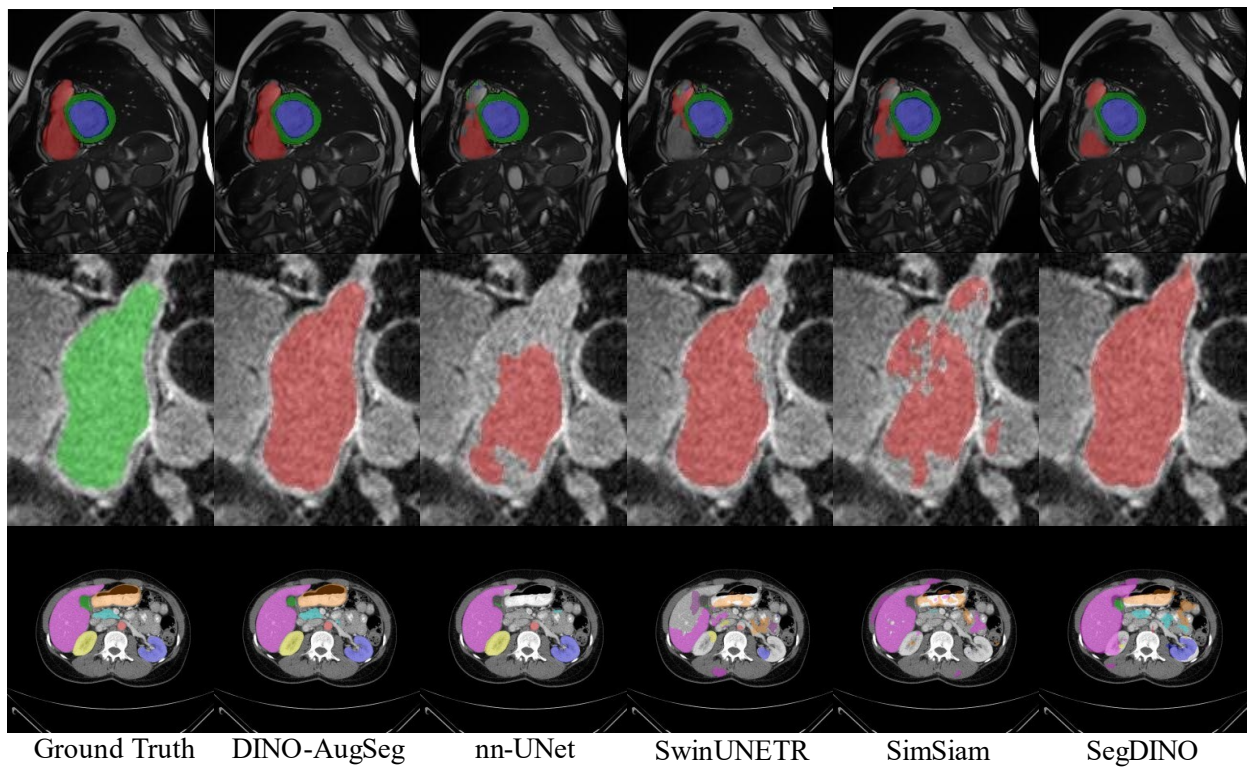

**Figure 8.** Representative qualitative segmentation results on the ACDC, LA2018, and Synapse datasets across MR (first and second rows) and CT (third row) modalities. The first column shows the ground-

truth overlay, while the remaining columns present the prediction results of different methods under the 7-shot training setting.

Figure 9 further presents a visual comparison under 25-shot, 10-shot, and 5-shot training settings on the TN3K, Kvasir-SEG, and ISIC2018 datasets (from top to bottom), showcasing representative segmentation results of the proposed DINO-AugSeg. The first column shows the ground-truth masks (green) overlaid on the original images, while the remaining columns present the predicted masks (blue) generated by DINO-AugSeg, nn-UNet, SwinUNETR, SimSiam, and SegDINO. Across all three datasets, DINO-AugSeg consistently achieves superior segmentation performance under few-shot learning scenarios. More qualitative examples under various few-shot settings across these five multi-modality datasets are provided in the supplementary material.

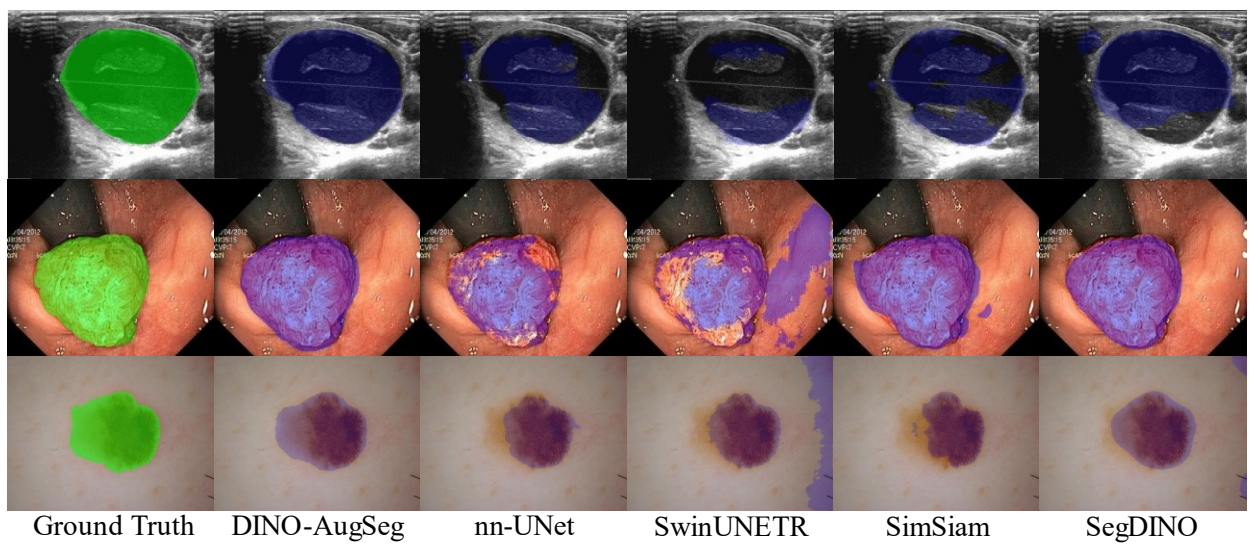

Ground Truth     DINO-AugSeg     nn-UNet     SwinUNETR     SimSiam     SegDINO

**Figure 9.** Representative segmentation results on the TN3K, Kvasir-SEG, and ISIC2018 datasets across ultrasound (US), endoscopy, and dermoscopy modalities under 25-, 10-, and 5-shot training settings, respectively. Ground-truth masks are shown in green, while model predictions are shown in blue.

## 5. Discussion

In this study, we exploit DINOv3-based features for few-shot medical image segmentation, focusing on feature-level augmentation and context-guided fusion. Experimental results across six datasets demonstrate the effectiveness of the proposed DINO-AugSeg framework. In this section, we highlight key insights and discuss limitations as potential directions for future research.

Data augmentation is a widely adopted strategy to increase the diversity of training samples, which is particularly important in medical image segmentation due to the limited availability of annotated data.

Unlike natural images, medical images often benefit from domain-specific augmentation strategies that integrate prior knowledge to improve model performance, such as atlas-based [27] and morphology-based methods [69]. In this work, we propose a wavelet-domain feature-level augmentation method for DINOv3 features. While this approach is effective for few-shot segmentation, a noticeable performance gap still remains compared to full-sample training. Incorporating domain priors as constraints in wavelet-based augmentation may be useful to further enhance performance. For instance, object-centric augmentation on DINOv3 features may be particularly beneficial, given the demonstrated capability of these features for object recognition and landmark detection.

Meanwhile, our results suggest that wavelet-domain feature-level augmentation may face some challenges in precise boundary delineation, especially under few-shot learning scenarios, as reflected by the HD95 metric. This indicates that shape-aware or structure-preserving augmentation in the wavelet domain may help guide the model to better capture the fine details of object contours. In future work, we aim to explore integrating object- and shape-related priors into wavelet-domain feature augmentation to further improve the robustness and boundary accuracy of few-shot medical image segmentation.

In addition, the decoder plays a critical role in integrating essential features and suppressing irrelevant information by progressively increasing the spatial resolution of feature maps through upsampling. Most existing approaches emphasize multi-scale feature fusion or assign importance weights across channels or spatial regions. In this study, we demonstrate that leveraging pre-trained DINOv3 features to guide decoder fusion through cross-attention is effective in few-shot learning scenarios. However, the advantage diminishes gradually as the number of training samples increases, when compared to state-of-the-art methods such as nn-UNet and SegFormer. This observation suggests that DINOv3 features, trained on natural images, may not be fully optimal for medical image analysis, reflecting the known domain gap between natural and medical images. A promising future direction lies in exploring advanced fine-tuning strategies, such as Adapters or Low-Rank Adaptation (LoRA) [8], to better adapt DINOv3 features to the medical domain. Furthermore, extending the proposed DINO-AugSeg framework to 3D object segmentation in MRI or CT data represents another valuable direction, as the current design is limited to 2D slice-based segmentation and does not explicitly exploit inter-slice spatial dependencies.

## 6. Conclusion

In this study, we explored the effectiveness of DINOv3 features for few-shot medical image segmentation. To address the challenges posed by limited training samples, we proposed WT-Aug, a wavelet-domain feature-level augmentation strategy, and CG-Fuse, a contextual information-guided feature fusion module for the decoder. Extensive experiments across six public benchmarks spanning five medical imaging modalities demonstrated that our framework, DINO-AugSeg, achieves strong performance under few-shot settings. These results highlight the effectiveness of WT-Aug and CG-Fuse in enhancing representation learning and improving segmentation accuracy, suggesting their potential as general strategies for advancing few-shot medical image segmentation.

## Acknowledgement

The study was supported by the US National Institutes of Health (R01 CA240808, R01 CA258987, R01 EB034691, and R01 CA280135).

segmentation', IEEE Transactions on Pattern Analysis and Machine Intelligence, 2025

45	Srivastava, N., Hinton, G., Krizhevsky, A., Sutskever, I., and Salakhutdinov, R.: 'Dropout: a simple way to prevent neural networks from overfitting', The journal of machine learning research, 2014, 15, (1), pp. 1929-1958

46	Chen, L.-C., Zhu, Y., Papandreou, G., Schroff, F., and Adam, H.: 'Encoder-decoder with atrous separable convolution for semantic image segmentation', in Editor (Ed.)^(Eds.): 'Book Encoder-decoder with atrous separable convolution for semantic image segmentation' (2018, edn.), pp. 801-818

47	Zhao, H., Shi, J., Qi, X., Wang, X., and Jia, J.: 'Pyramid scene parsing network', in Editor (Ed.)^(Eds.): 'Book Pyramid scene parsing network' (2017, edn.), pp. 2881-2890

48	Xie, E., Wang, W., Yu, Z., Anandkumar, A., Alvarez, J.M., and Luo, P.: 'SegFormer: Simple and efficient design for semantic segmentation with transformers', Advances in neural information processing systems, 2021, 34, pp. 12077-12090

49	Yang, S., Wang, H., Xing, Z., Chen, S., and Zhu, L.: 'Segdino: An efficient design for medical and natural image segmentation with dino-v3', arXiv preprint arXiv:2509.00833, 2025

50	Wu, Y., Li, X., Li, J., Yang, K., Zhu, P., and Zhang, S.: 'Dino is also a semantic guider: Exploiting class-aware affinity for weakly supervised semantic segmentation', in Editor (Ed.)^(Eds.): 'Book Dino is also a semantic guider: Exploiting class-aware affinity for weakly supervised semantic segmentation' (2024, edn.), pp. 1389-1397

51	Ayzenberg, L., Giryes, R., and Greenspan, H.: 'Dinov2 based self supervised learning for few shot medical image segmentation', in Editor (Ed.)^(Eds.): 'Book Dinov2 based self supervised learning for few shot medical image segmentation' (IEEE, 2024, edn.), pp. 1-5

52	Singh, P., Chukkapalli, R., Chaudhari, S., Chen, L., Chen, M., Pan, J., Smuda, C., and Cirrone, J.: 'Shifting to machine supervision: annotation-efficient semi and self-supervised learning for automatic medical image segmentation and classification', Scientific Reports, 2024, 14, (1), pp. 10820

53	Gao, Y., Li, H., Yuan, F., Wang, X., and Gao, X.: 'Dino u-net: Exploiting high-fidelity dense features from foundation models for medical image segmentation', arXiv preprint arXiv:2508.20909, 2025

54	Xu, G., Liao, W., Zhang, X., Li, C., He, X., and Wu, X.: 'Haar wavelet downsampling: A simple but effective downsampling module for semantic segmentation', Pattern recognition, 2023, 143, pp. 109819

55	Xiong, Z., Xia, Q., Hu, Z., Huang, N., Bian, C., Zheng, Y., Vesal, S., Ravikumar, N., Maier, A., and Yang, X.: 'A global benchmark of algorithms for segmenting the left atrium from late gadolinium-enhanced cardiac magnetic resonance imaging', Medical image analysis, 2021, 67, pp. 101832

56	Yu, L., Wang, S., Li, X., Fu, C.-W., and Heng, P.-A.: 'Uncertainty-aware self-ensembling model for semi-supervised 3D left atrium segmentation', in Editor (Ed.)^(Eds.): 'Book Uncertainty-aware self-ensembling model for semi-supervised 3D left atrium segmentation' (Springer, 2019, edn.), pp. 605-613

57	Gong, H., Chen, J., Chen, G., Li, H., Li, G., and Chen, F.: 'Thyroid region prior guided attention for ultrasound segmentation of thyroid nodules', Computers in biology and medicine, 2023, 155, pp. 106389

58	Jha, D., Smedsrud, P.H., Riegler, M.A., Johansen, D., De Lange, T., Halvorsen, P., and Johansen, H.D.: 'Resunet++: An advanced architecture for medical image segmentation', in Editor (Ed.)^(Eds.): 'Book Resunet++: An advanced architecture for medical image segmentation' (IEEE, 2019, edn.), pp. 225-2255

59	Codella, N., Rotemberg, V., Tschandl, P., Celebi, M.E., Dusza, S., Gutman, D., Helba, B., Kalloo, A., Liopyris, K., and Marchetti, M.: 'Skin lesion analysis toward melanoma detection 2018: A challenge hosted by the international skin imaging collaboration (isic)', arXiv preprint arXiv:1902.03368, 2019

60	Schlemper, J., Oktay, O., Schaap, M., Heinrich, M., Kainz, B., Glocker, B., and Rueckert, D.: 'Attention gated networks: Learning to leverage salient regions in medical images', Medical image analysis, 2019, 53, pp. 197-207

61	Ibtehaz, N., and Rahman, M.S.: 'MultiResUNet: Rethinking the U-Net architecture for multimodal biomedical image segmentation', Neural networks, 2020, 121, pp. 74-87

62	Badrinarayanan, V., Kendall, A., and Cipolla, R.: 'Segnet: A deep convolutional encoder-decoder
28

# Supplementary material

# Exploiting DINOv3-Based Self-Supervised Features for Robust Few-Shot Medical Image Segmentation

**1. Results on the ACDC, LA2018, and Synapse Datasets (Volumetric Image) with Two Training Scans**

We conducted experiments on the ACDC, LA2018, and Synapse datasets using only two training scans to evaluate the effectiveness of the proposed DINO-AugSeg framework under extremely limited supervision. Table 1 presents the Dice and HD95 metrics across a broad range of segmentation models, including convolution-based methods (Conv), transformer-based architectures (Trans), and self-supervised learning approaches (SSL).

Overall, DINO-AugSeg obtains the highest Dice on ACDC (76.24) and LA2018 (81.01) and yields competitive performance on Synapse (50.37), where MoCov2 (51.91) achieves the top Dice score. In terms of boundary accuracy (HD95), DINO-AugSeg attains very low HD95 on ACDC (15.56) and LA2018 (23.57), while on Synapse its HD95 (37.25) is the second best (AIMv2: 36.93).

**Table 1.** Experimental results on the ACDC, LA2018, and Synapse datasets (volumetric image) with *two* training scans. Arrows are pointing in the direction of improved accuracy. The bold metric results indicate the best values.

| Dataset | | ACDC | | LA2018 | | Synapse | |
|---|---|---|---|---|---|---|---|
| Method | Type | DICE (%)↑ | HD95 (vol)↓ | DICE (%)↑ | HD95 (vol)↓ | DICE (%)↑ | HD95 (vol)↓ |
| Attention-Unet | Conv | 40.10 | 72.50 | 50.51 | 45.16 | 49.94 | 64.08 |
| MultiResUNet | Conv | 7.64 | 105.56 | 34.44 | 49.79 | 5.410 | 221.41 |
| nn-UNet | Conv | 27.47 | 40.09 | 71.39 | 24.89 | 40.24 | 108.11 |
| SegNet | Conv | 30.04 | 76.89 | 61.14 | 32.20 | 24.10 | 55.65 |
| SegResNet | Conv | 19.17 | 51.64 | 53.04 | 41.25 | 38.09 | 74.09 |
| U-Net | Conv | 31.93 | 47.45 | 51.86 | 38.41 | 46.35 | 44.35 |
| Unet++ | Conv | 32.77 | 60.33 | 45.44 | 39.64 | 49.82 | 41.73 |
| MISSFormer | Trans | 33.46 | 26.45 | 66.38 | 20.68 | 27.16 | 66.28 |
| SegFormer | Trans | 11.62 | 66.39 | 52.95 | 29.70 | 22.80 | 72.55 |
| SwinUNETR | Trans | 32.17 | 56.37 | 54.25 | 33.48 | 29.47 | 60.36 |
| AIMv2 | SSL | 51.52 | 25.93 | 61.09 | 19.64 | 38.77 | **36.93** |
| MAE | SSL | 36.91 | 16.49 | 57.50 | 27.41 | 32.65 | 63.82 |
| MoCov2 | SSL | 58.62 | 31.33 | 66.38 | **20.68** | **51.91** | 39.89 |
| SimSiam | SSL | 53.06 | 32.00 | 49.01 | 36.08 | 47.00 | 46.72 |
| SegDINO | SSL | 57.13 | 28.63 | 62.24 | 30.53 | 38.30 | 41.38 |
| DINO-AugSeg | SSL | **76.24** | **15.56** | **81.01** | 23.57 | 50.37 | 37.25 |

## 2. Results on the ACDC, LA2018, and Kvasir-SEG Datasets with Full Training Scans

We further evaluated all competing methods under the full-data setting, where all available training scans are used. This experiment aims to assess the upper-bound performance of each method when data scarcity is not a limiting factor.

Table 2 summarizes the results on the ACDC, LA2018, and Kvasir-SEG datasets. Classical convolution-based models such as Attention-UNet, SegNet, and SegResNet demonstrate solid performance on ACDC and LA2018 but struggle on the more challenging Kvasir-SEG dataset, particularly in terms of boundary accuracy as reflected by the poor HD95 values. Transformer-based models (MISSFormer, SegFormer, and SwinUNETR) achieve competitive Dice scores on ACDC and LA2018, while exhibiting mixed performance on Kvasir-SEG. Self-supervised learning (SSL) methods show notable improvements, with SegDINO achieving strong overall results across datasets. Our proposed DINO-AugSeg achieves the highest Dice score on ACDC (91.29%), LA2018 (90.88%), and Kvasir-SEG (86.18%). Importantly, it achieves the lowest HD95 (59.96) on Kvasir-SEG by a large margin, indicating superior boundary fidelity in complex endoscopy scenes.

These results demonstrate that DINO-AugSeg effectively leverages augmentation-enhanced representation learning to boost segmentation performance, even when training data is abundant. Its consistent improvements across both Dice and HD95 metrics highlight its strong generalization capability and robustness across diverse imaging modalities. However, when comparing these full-data results with the few-shot setting, we observe that the advantage of DINO-AugSeg becomes less pronounced. In particular, on some datasets, its improvement over strong baselines such as nn-UNet and SegDINO is relatively modest under the full-training regime. This suggests that the primary benefit of DINO-AugSeg lies in enhancing representation quality when data is limited, while under abundant data conditions, the performance gap among methods narrows.

**Table 2.** Experimental results with full training samples on ACDC, LA2018 and Kvasir-SEG. Arrows are pointing in the direction of improved accuracy. The bold metric results indicate the best values.

| Dataset | | ACDC | | LA2018 | | Kvasir-SEG | |
|---|---|---|---|---|---|---|---|
| Method | Type | DICE (%)↑ | HD95 (vol)↓ | DICE (%)↑ | HD95 (vol)↓ | DICE (%)↑ | HD95 (vol)↓ |
| Attention-Unet | Conv | 91.04 | 2.47 | 71.15 | 28.49 | 77.65 | 124.13 |
| MultiResUNet | Conv | 60.88 | 37.19 | 50.91 | 48.31 | 31.50 | 221.85 |
| nn-UNet | Conv | 91.16 | 1.65 | 90.64 | **5.39** | 72.92 | 114.35 |
| SegNet | Conv | 90.14 | 4.57 | 86.29 | 17.53 | 66.93 | 125.70 |
| SegResNet | Conv | 89.88 | 1.82 | 84.15 | 20.20 | 78.04 | 123.71 |
| U-Net | Conv | 88.81 | 12.88 | 79.37 | 18.00 | 75.10 | 134.76 |
| Unet++ | Conv | 90.74 | 4.82 | 83.20 | 18.37 | 73.63 | 139.57 |
| MISSFormer | Trans | 88.57 | **1.56** | 90.35 | 6.64 | 75.76 | 105.99 |
| SegFormer | Trans | 88.86 | 6.64 | 89.01 | 12.04 | 65.85 | 117.26 |
| SwinUNETR | Trans | 88.37 | 2.86 | 89.55 | 11.53 | 71.80 | 108.45 |
| AIMv2 | SSL | 76.96 | 4.11 | 87.86 | 12.37 | 72.79 | 107.00 |
| MAE | SSL | 75.69 | 7.55 | 78.62 | 19.25 | 64.20 | 163.17 |
| MoCov2 | SSL | 90.62 | 3.13 | 78.84 | 23.72 | 81.81 | 110.19 |
| SimSiam | SSL | 90.13 | 2.35 | 75.62 | 21.28 | 79.89 | 125.49 |
| SegDINO | SSL | 91.04 | 1.72 | 90.48 | 5.98 | 73.57 | 98.20 |
| DINO-AugSeg | SSL | **91.29** | 1.66 | **90.88** | 6.32 | **86.18** | **59.96** |

## 3. Effect of Model Scale on DINO-AugSeg

We further examined the impact of model scale on the proposed DINO-AugSeg framework by evaluating three configurations: Small, Base, and Large, under the 7-shot training setting. Experiments were conducted on the ACDC and LA2018 datasets to assess whether increasing model capacity could enhance segmentation performance in low-data regimes.

As summarized in Table 3, segmentation performance improved consistently as the model size increased. The Large model achieved the best overall results, with Dice scores of 81.85% and 86.22%, and HD95 values of 9.96 and 11.75 on ACDC and LA2018, respectively. These gains suggest that scaling up the DINOv3 backbone enables the model to better capture subtle anatomical structures and complex spatial variations, ultimately improving robustness across datasets.

**Table 3.** Segmentation performance of DINO-AugSeg with different model sizes (Small, Base, Large) under the 7-shot training setting on the ACDC and LA2018 datasets. Arrows are pointing in the direction of improved accuracy.

| Dataset | ACDC | | LA2018 | |
|---|---|---|---|---|
| Model Size | DICE(%)↑ | HD95(vol)↓ | DICE(%)↑ | HD95(vol)↓ |
| Small | 77.59 | 19.33 | 81.73 | 24.02 |
| Base | 80.72 | 10.22 | 83.75 | 20.67 |
| Large | 81.85 | 9.96 | 86.22 | 11.75 |

## 4. Statistics Analysis

To further assess the significance of performance differences among the compared methods, we conducted pairwise statistical testing using the Wilcoxon signed-rank test. Specifically, p-values were computed between each baseline method and the proposed DINO-AugSeg model. This non-parametric test evaluates whether the performance improvements introduced by DINO-AugSeg are statistically significant across all test samples. We applied the Wilcoxon test to the Dice score of each segmentation target.

As shown in Table 4, DINO-AugSeg achieves statistically significant improvements over most competing methods, with p-values below 0.05 across anatomical structures (RV= right ventricle, MYO= myocardium, LV= left ventricle) under both one-shot and seven-shot settings. The only exceptions are SegDINO, SimSiam, and MoCoV2, which exhibit p-values greater than 0.05 for one or two structures.

This observation suggests that these three approaches—each based on self-supervised pretrained encoders—may produce feature representations that are more similar to those learned by DINOv3. Consequently, the performance differences between these SSL-based methods and DINO-AugSeg are smaller and therefore may not reach statistical significance in certain cases. This observation is further supported by the results on four additional datasets (ISIC2018, Kvasir-SEG, TN3K, and LA2018), as shown in Table 5. On these datasets, SSL-based methods such as AIMv2, MoCov2, and SegDINO also exhibit p-values greater than 0.05 in several settings.

**Table 4.** Wilcoxon signed-rank statistical tests evaluating p-values on the ACDC dataset under one-shot and seven-shot training settings.

| Method | 1 Training Sample | | | 7 Training Samples | | |
|---|---|---|---|---|---|---|
| Objects | RV | MYO | LV | RV | MYO | LV |
| Attention-Unet | < 0.05 | < 0.05 | < 0.05 | < 0.05 | < 0.05 | < 0.05 |
| MultiResUNet | < 0.05 | < 0.05 | < 0.05 | < 0.05 | < 0.05 | < 0.05 |
| nn-UNet | < 0.05 | < 0.05 | < 0.05 | < 0.05 | < 0.05 | < 0.05 |
| SegNet | < 0.05 | < 0.05 | < 0.05 | < 0.05 | < 0.05 | < 0.05 |
| SegResNet | < 0.05 | < 0.05 | < 0.05 | < 0.05 | < 0.05 | < 0.05 |
| U-Net | < 0.05 | < 0.05 | < 0.05 | < 0.05 | < 0.05 | < 0.05 |
| Unet++ | < 0.05 | < 0.05 | < 0.05 | < 0.05 | < 0.05 | < 0.05 |
| MISSFormer | < 0.05 | < 0.05 | < 0.05 | < 0.05 | < 0.05 | < 0.05 |
| SegFormer | < 0.05 | < 0.05 | < 0.05 | < 0.05 | < 0.05 | < 0.05 |
| SwinUNETR | < 0.05 | < 0.05 | < 0.05 | < 0.05 | < 0.05 | < 0.05 |
| AIMv2 | < 0.05 | < 0.05 | < 0.05 | < 0.05 | < 0.05 | < 0.05 |
| MAE | < 0.05 | < 0.05 | < 0.05 | < 0.05 | < 0.05 | < 0.05 |
| MoCov2 | < 0.05 | < 0.05 | < 0.05 | < 0.05 | < 0.05 | **0.09** |
| SimSiam | < 0.05 | < 0.05 | < 0.05 | < 0.05 | < 0.05 | **0.65** |
| SegDINO | **0.07** | < 0.05 | < 0.05 | < 0.05 | < 0.05 | **0.31** |

**Table 5.** Wilcoxon signed-rank statistical tests evaluating p-values on the ISIC2018, Kvasir-SEG, TN3K, and LA2018 datasets under few-shot training settings.

| Dataset | ISIC2018 | | Kvasir-SEG | | TN3K | | LA2018 | |
|---|---|---|---|---|---|---|---|---|
| Method | 5 Training Samples | 25 Training Samples | 10 Training Samples | 40 Training Samples | 25 Training Samples | 100 Training Samples | 1 Training Sample | 7 Training Samples |
| Attention-Unet | < 0.05 | 0.06 | < 0.05 | < 0.05 | < 0.05 | < 0.05 | < 0.05 | < 0.05 |
| MultiResUNet | < 0.05 | < 0.05 | < 0.05 | < 0.05 | < 0.05 | < 0.05 | < 0.05 | < 0.05 |
| nn-UNet | < 0.05 | < 0.05 | < 0.05 | < 0.05 | < 0.05 | < 0.05 | < 0.05 | < 0.05 |
| SegNet | < 0.05 | < 0.05 | < 0.05 | < 0.05 | < 0.05 | < 0.05 | < 0.05 | < 0.05 |
| SegResNet | < 0.05 | 0.06 | < 0.05 | < 0.05 | < 0.05 | < 0.05 | < 0.05 | < 0.05 |
| U-Net | < 0.05 | < 0.05 | < 0.05 | < 0.05 | < 0.05 | < 0.05 | < 0.05 | < 0.05 |
| Unet++ | < 0.05 | 0.71 | < 0.05 | < 0.05 | < 0.05 | < 0.05 | < 0.05 | < 0.05 |
| MISSFormer | < 0.05 | 0.99 | < 0.05 | < 0.05 | < 0.05 | < 0.05 | < 0.05 | < 0.05 |
| SegFormer | < 0.05 | < 0.05 | < 0.05 | < 0.05 | < 0.05 | < 0.05 | < 0.05 | < 0.05 |
| SwinUNETR | < 0.05 | < 0.05 | < 0.05 | < 0.05 | < 0.05 | < 0.05 | < 0.05 | < 0.05 |
| AIMv2 | **0.26** | < 0.05 | < 0.05 | < 0.05 | < 0.05 | < 0.05 | < 0.05 | < 0.05 |
| MAE | < 0.05 | < 0.05 | < 0.05 | < 0.05 | < 0.05 | < 0.05 | < 0.05 | < 0.05 |
| MoCov2 | **0.07** | < 0.05 | < 0.05 | < 0.05 | < 0.05 | < 0.05 | < 0.05 | < 0.05 |
| SimSiam | < 0.05 | < 0.05 | < 0.05 | < 0.05 | < 0.05 | < 0.05 | < 0.05 | < 0.05 |
| SegDINO | **0.85** | < 0.05 | < 0.05 | < 0.05 | < 0.05 | **0.14** | < 0.05 | < 0.05 |

## 5. Visual Comparison

(1) Few-shot results visualization on the ACDC, LA2018, and Synapse datasets

Figure 1 illustrates representative segmentation results on the ACDC, LA2018, and Synapse datasets across MR (first and second rows) and CT (third row) modalities under the 1-shot training setting. The first column shows the ground-truth masks overlaid on the original images, while the remaining columns present the predictions generated by DINO-AugSeg, nn-UNet, SwinUNETR, SimSiam, and SegDINO.

...

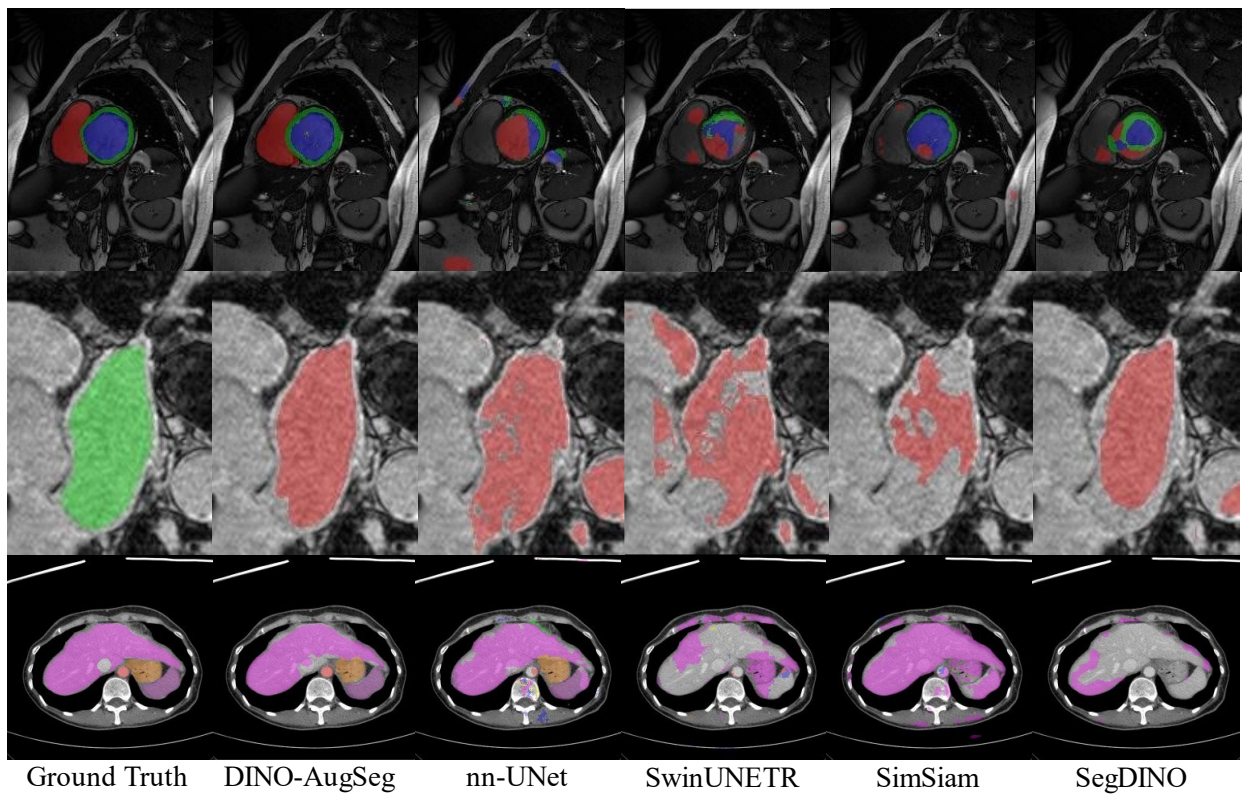

**Figure 1.** Representative segmentation results on the ACDC, LA2018, and Synapse datasets across MR (first and second rows) and CT (third row) modalities. The first column shows the ground-truth overlay, while the remaining columns present the prediction results under the 1-shot training setting.

(2) Few-shot results visualization on the TN3K, Kvasir-SEG, and ISIC2018 datasets

Figure 2 shows a qualitative comparison under the 100-shot, 40-shot, and 25-shot training settings on the TN3K, Kvasir-SEG, and ISIC2018 datasets (top to bottom), highlighting representative segmentation results of the proposed DINO-AugSeg. The first column displays the ground-truth masks (green) overlaid on the corresponding images, while the following columns present predicted masks (blue) from DINO-AugSeg, nn-UNet, SwinUNETR, SimSiam, and SegDINO. Across all datasets and few-shot settings, DINO-AugSeg consistently achieves superior segmentation performance.

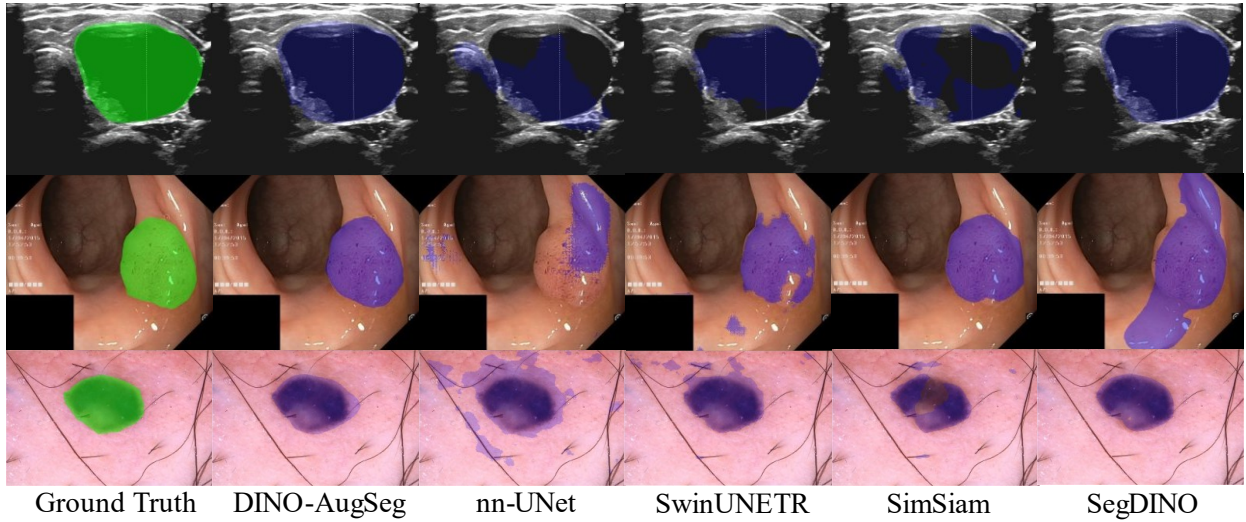

Ground Truth   DINO-AugSeg   nn-UNet   SwinUNETR   SimSiam   SegDINO

**Figure 2.** Representative segmentation results on the TN3K, Kvasir-SEG, and ISIC2018 datasets across ultrasound (US), endoscopy, and dermoscopy modalities under 100-, 40-, and 25-shot training settings. Ground-truth masks are shown in green, while model predictions are shown in blue.